# Medical Images Analysis in Cancer Diagnostic

Jelena Vasiljević[1], Ivica Milosavljević[2], Vladimir Krstić[1], Nataša Zivić[3], Lazar Berbakov[1], Luka Lopušina[5], Dhinaharan Nagamalai[4] and Milutin Cerović[5]

[1]Institute Mihajlo Pupin, University of Belgrade, Belgrade, Serbia
[2]Center for Pathology and Forensics of the Military Medical Academy in Belgrade, Belgrade, Serbia
[3]University of Siegen, Germany
[4]Wireill, Australia
[5]The School of Computing, University Union

*ABSTRACT*

*This paper shows results of computer analysis of images in the purpose of finding differences between medical images in order of their classifications in terms of separation malign tissue from normal and benign tissue. The diagnostics of malign tissue is of the crucial importance in medicine. Therefore, ascertainment of the correlation between multifractals parameters and "chaotic" cells could be of the great appliance. This paper shows the application of multifractal analysis for additional help in cancer diagnosis, as well as diminishing. of the subjective factor and error probability.*

## 1. Application of Multifractal Analysis in Image Processing

Multifractal analysis of images is based on the definition of measurements with images that are gray levels. Then the multifractal spectrum is calculated. In contrast to many classic approaches, there is no filtering. The spectrum uses local as well as global information for segmentation, noise reduction or edge detection at image points.

Image analysis is a fundamental component of a computer visual problem, with applications in robotics, medical or satellite images ... Segmentation is an important step that provides a description of a basic individual process. Filtering then gives signal gradients where extremes roughly correspond to contours. Then, multi-resolution techniques can be used to "purify" the results obtained. The main drawback of this approach is loss in precision due to preliminary filtering.

An alternative approach is to observe that a picture is a measure known as a fixed resolution. The irregularities of this measure can then be studied using a multifractal analysis. The general principle is the following: first, the different dimensions and capacitance are defined from the image which is the gray level. Then, the corresponding multifractal spectrum is calculated, providing both local (over) and global (through) information. There are no hypotheses about the regularity of the signal. Multifractal analysis (MF) can be successfully used in image processing. The idea of applying (inverse) MF analysis in extracting characteristic details in the picture is presented in [1].

The importance and the advantage of fractal and multifactal analysis (MFA), in relation to the "classic" signal analysis, lies in the way in which irregularity is considered. The MFA tries to extract information directly from singularity, while in the "classic" mode, most often, NF filtered versions are viewed, possibly with different filtering depths, to detect irregularities and suppress noise. In particular, based on a certain value i, the points of inhomogeneity in the original signal can be separated [1, 2, 3, 4]. By dividing pixel images that satisfy the selected parameter value, or spectrum, by inverse multifractal analysis (IMFA), it is possible to extract from the image of a region that can not otherwise be noticed by any of the known methods. An additional advantage is that such a segmentation does not cause any degradation of the initial image: all the mutual relations of the pixels remain unchanged, so that the details of the image are kept completely. This feature is particularly important in medical diagnostics, so the potential of IMFA in this area is high.

It is shown that a large number of frequently variables of a different nature (electrical signals, modern telecommunication traffic, meteorological and biomedical signals) can be described in a similar way. It is necessary to examine the fractal characteristics for the expression of significant variability. The use of classical statistical methods in such a case (mean value) could cause error estimates. The pronounced singularities indicate the multifunctionality of the process.

## 2. FRACTAL MORPHOMETRY APPLIED TO TUMORS

Despite the huge increase in our understanding of the molecular carcinoma mechanism, most diagnoses are still determined by visual examination of radiographic images, microscopic and biopsy patterns, direct examination of the tissue, and so on. Usually, these techniques are applied in a quality manner by clinicians who are trained to classify images showing abnormalities such as structural irregularities or high indications for mitosis. A more qualitative and reproducible method, which can serve as an ancillary tool for diagnostic training, is an analysis of images using computer tools. This lies in the potential of fractal analysis as a morphometric measure of irregular structures that are typical of tumor growth.

Pathologists are skilled in examining the boundary surface of the epithelial-connective tissue, which separates the tumor and surrounding healthy tissue. The nature of the tumor edge, whether infiltrative or invasive or poorly expansive, provides information useful not only for prognosis, but also for diagnosis (either benign or malignant tumors). In the study of Landini and Ripini [5], the border area of the epithelial-connective tissue of oral mucosa was examined. Lesions are classified by routine diagnosis into four categories: a) normal; b) medium dysplasia; c) moderate to severe dysplasia; d) carcinomas. Fractal lesion analysis, which subsequently followed, revealed the following fractal dimensions for the above four categories: 1.07 ± 0.05, 1.08 ± 0.09, 1.16 ± 0.08, and 1.41 ± 0.08, respectively. Although the differences were not large enough to be accepted as an independent tool in diagnostics, they are regardless of consistent measurements of the degree of distortion of the boundary surface. Landini and Ripini then proceeded to describe using a more sophisticated multifunctional analysis method that gives a spectrum of fractal values instead of one value for each image. This method has provided more reliable discrimination of the pathological conditions of the tissue. Lefebre and Benali [6] and Polman et al [7] have shown that fractal methods can also be useful for analyzing digitized mammograms, increasing the hope that the number of incorrect positive mammograms will be reduced in this way. Considering that the increase in irregularities, with associated fractal fracture enlargement, is a common indicator of tumor growth this undoubtedly represents a universal result. Out of everything contained in this chapter Fractals in biomedical systems, the hypotheses set out in this paper are derived.

## 2.1. Histopathological characteristics of normal mucous membrane, adenoma and adenocarcinoma of the colon

The normal mucous membrane of the mucous membrane of the colon is from the lumen of the intestine to the surface of the laminae epithelialis tunicae mucosae consisting of a layer of cylindrical goblet cells, absortive, endocrine and undifferentiated cells lying on the basal membrane and invaginating the so-called crypts They reach the next layer, laminae muscularis mucosae built of smooth muscle cells, and underneath there is a lamina propria of tuniciae mucosae with blood vessels, loose connective tissue, and individual mesenchymal cells.

The basic histological characteristics that point to the normality, regularity are: uniformity of the cells of the liminal epithelialis tunicae mucosae, polarization of the sails to the basal memebran, small sails, preserved secretion of the mucin by the goblet cells. Figure 1 shows a photo of a sample of the normal colon tissue, he (hematoxilieosin-coloring method), a transverse cross sectional view, and in Fig. 2, photographs of a sample of the normal tissue of the column, he is a longitudinal sectional view.

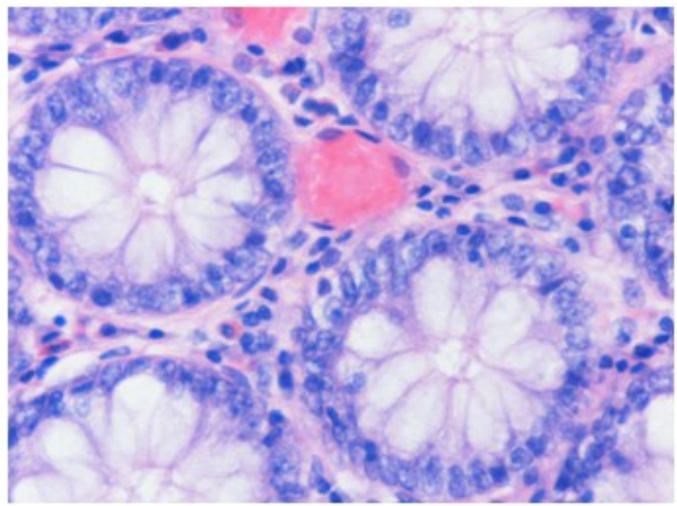

Figure 1: Photo of a sample of normal colon tissue, (he), a transversal section view

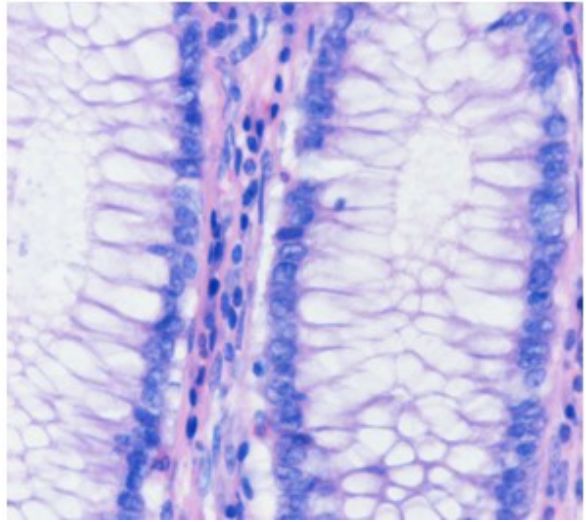

Figure 2: Photo of a sample of normal colon tissue, (he), a longitudinal section view

The most common tumors in the colon are the origin of the epithelium and they can be benign-nature-adenomas and malignant-nature-adenocarcinomas. Adenoma is a benign tumor of the epithelial origin in which proliferation of the epithelium occurs with the formation of glandular structures, which are coated with dysplastic cells. The disposable epithelium loses the uniformity of its cells, the cells are pleomorphic, and the usual architectural appearance is lost, tubular and / or vilosic structures are formed which are coated with cylindrical epithelial cells, with elongated and hyperchromatic nuclei and which may and may not have preserved mucigen activity, while the mitotic activity of the cells is increased. In Figure 3, a photo of the sample of the adenoma column is shown, he is a transversal sectional view, and in Figure 4 a photo of the sample of the adenoma column is shown, and he shows the longitudinal cross section.

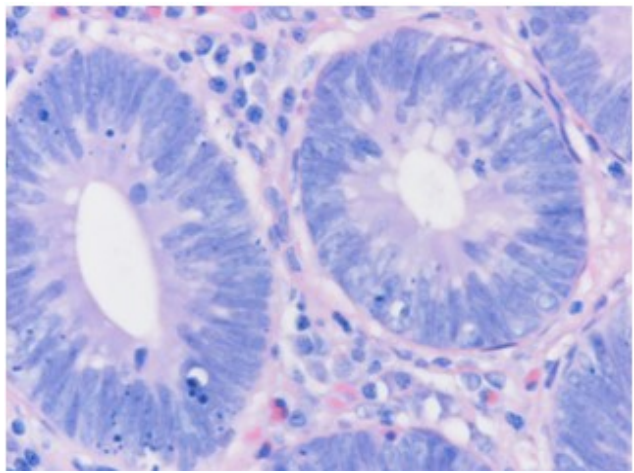

Figure 3: Photo of the column adenoma sample, (he), a transverse cross sectional view

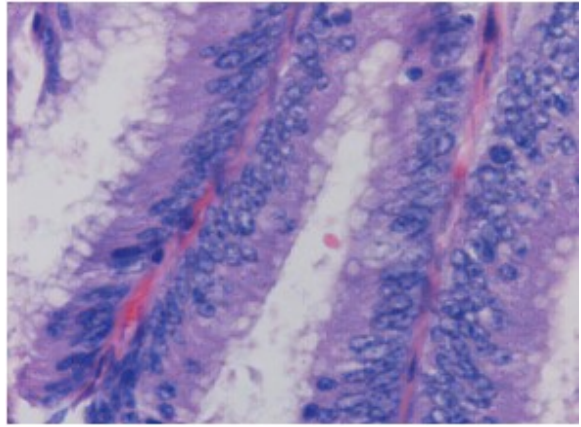

Figure 4: Photo of the column adenoma sample, (he), a longitudinal section view

Adenocarcinoma of the colon is one of the most common tumors in the human population and is one of the major challenges of human medicine, precisely because it can arise from the adenomas and what produces the symptoms relatively early, which allows diagnosis and treatment. Carcinoma tissue is constructed of tubular formations, irregular shape and size, as well as from cribriform formations, which are coated with cubic and cylindrical atypical cells whose nuclei are pleomorphic, hyperchromatic, with and without prominent cats. The surrounding stroma of the tumor is desmoplastic, with multiplied binders and with inflammatory lymphocyte infiltrate, plasmocyte, and granulocyte. The properties of adenocarcinoma are: local invasive growth by penetrating the basal membrane, infiltrating the column wall and possibly spreading into the surrounding structures, as well as the ability to metastasize to regional lymph nodes, and metastasis to distant organs: in the liver, lungs, bones, and other organs [ CECI89]. Figure 5 shows a photograph of a sample of colon carcinoma, a he-transversal sectional view, and Figure 6 shows a photo of a sample of colon carcinoma, he- an indication of the longitudinal cross-section.

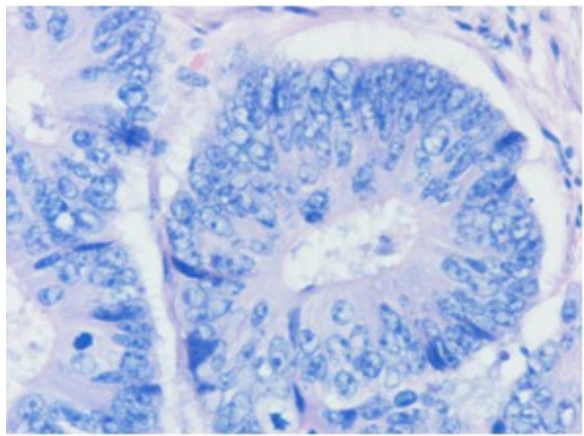

Figure 5: Photo of the column carcinoma sample, (he), a transverse cross-sectional view

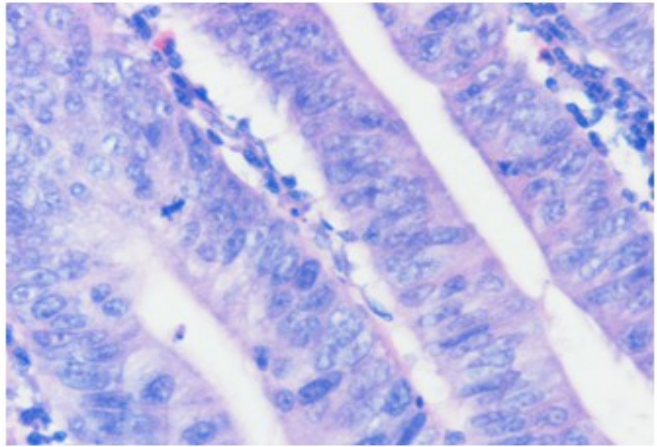

Figure 6: Photo of the column carcinoma sample, (he), a longitudinal section view

## 2.2. Methodological part of the research

**Research goals**

The intention behind this research is to determine the existence of differences in the parameters of the multifractal analysis of digital medical images between the following three tissue groups:

1. Normal mucosal tissue of the colon

2. Thick bowel mucosal tissue with diagnosis of malignant tumor origin of the epithelium – carcinomas

3. Thick bowel mucosal tissue diagnosed as benign tumors of the origin of the epithelium – adenomas

In order to obtain the required research results, it is necessary to perform the following steps:

1. Determine the parameters of the multifaceted analysis for the previously stated groups of tissue pictures

2. Determine the existence of statistically significant differences between the parameters corresponding to the previously mentioned tissue image groups.

## 2.3. Method

It is a non-experimental correlation study on the sample. In relation to the goal, the research is parametric.

## 2.4. Variables of research

Independent variable

1. Type of photographed tissue. Variables are operatively defined with the following categories:

a) tissue without pathological changes

b) tissue with cancer

c) adenoma tissue

Dependent variables

In the FracLac program

1. $D_{max}$ - maximum

2. $\overline{Q}$ - Q which corresponds to the maximum

3. $\underline{\alpha}$ - which corresponds to the minimum

4. $f(\alpha)_{min}$ - minimum

5. $\overline{\alpha}$ - which corresponds to the maximum

6. $f(\alpha)_{max}$ - maximum

In FracLab

1. $\alpha_{sr}$ - mean value

2. $f(\alpha)_{sr}$ - mean value

3. $\overline{\alpha}$ - which corresponds to the maximum

4. $\alpha_{stdev}$ - standard deviation

5. $f(\alpha)_{stdev}$ - standard deviation

## 2.5. Instruments

The following programs were used for multifractal analysis of the obtained digital medical images and obtaining the parameters of multifactal analysis:

1. "FracLac" program for multifractal image analysis [8].

2. "FracLab" program for multifractal image analysis [9].

3. Program for statistical analysis of data SPSS (Statistical Package for Social Sciences), standard for statistical analysis of clinical research results.

## 2.6. FracLac

The "FracLac" program is created in the Java programming language and represents one of the plugins in the "ImageJ" digital image analysis program. "ImageJ" is freely available image analysis software written in the Java programming language authored by Wayne Rasband of the National Institute of Health of the United States of America from Bethesda, Maryland, [8].

The author of "FracLac" is Audrey Karperian from Charles Sturt University in Australia. She was contacted with this research. At her request, they were sent one sample sample from each of the three groups to test and improve the program for the specific problem of this research. With great help from her professional team, this program is adapted to handle a large number of images at once, relatively quickly. This is especially suited for use by medical personnel, where it would not be complicated training, nor would it take up a lot of their time. The images necessary for processing can be collected and then all at once processed.

## 2.7. The program "FracLab"

The FracLab program was developed by a team of experts at the INRIA and IRCCyN Institutes in France, led by Jacques Levy Vehell.

The principle of image segmentation using multifactal analysis is as follows: the points lying in the picture can be classified according to their Holder exponent. Let's look at the example of the points that lie on the contours. These points often correspond to the discontinuities of the gray level map or from its output. They therefore generally have a "low" Holder regularity. However, the exact value of the exponent will depend on the characteristics of the image. Additionally, the boundary edge feature is not purely local, and therefore a global criterion is required to decide whether a point is assigned, a point that belongs to the edge. Indeed, the points lying on the textures of the region also have a generally low regularity, and it is necessary to find a way to make them different from contours. Here, the other component of the multifractal analysis comes to the fore: since the edges are by definition the sets of points of the dimension one, we declare that the point lies on the contour if there is an exponent such that the associated value of the multifractal spectrum is one. In addition to the geometric characterization of the edges of the edges, it is also possible to make statistical points: the points of the edges can be defined by their probability of being affected when the pixel is randomly selected in the image at a given resolution. The relationship between the geometric and statistical representation of the edges of the edge provides multifractal formalism. Instead of edge detection, a much more complicated structure can be extracted using the same principle: starting again from Holder's exponents, points can be kept where the spectrum has a certain value. For example, by selecting a value of about 1.5, it is generally possible to extract very irregular contours. The value close to 2 corresponds either with smooth regions or textures.

The general procedure is as follows: it begins by calculating the Holder exponent at each point. This gives a picture of Holder's exponents.

The second step is calculating the multifunctional spectrum. In this paper the Hausdorf spectrum is calculated from the three spectra offered. The Hausdorf spectrum gives geometric information that relates to the dimension of the set of points in the image with the given exponent. This spectrum is a function in which apscis represents all Holder's exponents appearing in the image, and the ordinate is the dimension of the set of pixels with the given exponent.

The second spectrum is a large deviation spectrum that gives statistical information related to the probability of finding a point with a given exponent in the image (or more precisely, as this probability acts in the change of resolution).

The third spectrum is the so-called. The Leandre's spectrum, which represents only the concave approximation of the spectrum of large deviations, and its main contribution is to give much more robust calculations, although at the cost of losing information.

## 2.8. Hypotheses

General hypotheses

The parameters of multifractal analysis will significantly differ from all three groups: normal tissue, carcinomas and adenomas.

1. In particular, in the case of multifractal analysis of digital images of three groups observed, FracLac will distinguish the following obtained parameters for all three groups observed:

- $D_{max}$ - maximum

- $\overline{Q}$ - Q that corresponds to the maximum

- $\underline{\alpha}$ - which corresponds to the minimum

- $f(\alpha)_{min}$ - minimum

- $\overline{\alpha}$ - which corresponds to the maximum

- $f(\alpha)_{max}$ - maximum

2. In the case of multifractal analysis of digital images of the three groups observed, FracLab will distinguish the following obtained parameters for all three groups observed:

- $\alpha_{sr}$ - Middle value

- $f(\alpha)_{sr}$ - Middle value

- $\overline{\alpha}$ - which corresponds to the maximum

- $\alpha_{stdev}$ - standard deviation

- $f(\alpha)_{stdev}$ - standard deviation

**2.9. Sample**

The sample consisted of 150 preparations obtained from biopsy from the gastrointestinal tract, more precisely from the colon, Adenocarcinoma tubulare coli. Of the 150 preparations, 50 were previously diagnosed as normal colon mucosal tissue, 50 were diagnosed as colon mucosal tissue with malignant epithelial tumors - carcinomas, and 50 diagnosed as colon mucosal tissue with benign epithelial tumors - adenomas.

In the Center for Pathology and Forensics of the Military Medical Academy in Belgrade, the preparations were prepared for analysis under a microscope with a magnification of 40x and photographed on a coolscope device (by the author with the help of Dr. Ivana Tufegčić) in digital form (Figures 1, 2 and 3). Coolscope is a kind of hybrid microscope in the body of the computer, the manufacturer is a Japanese firm, Nikon. Five different photographs were taken from each preparation, in order to obtain the most valid results of statistical analysis. In this way, a total of 750 digital images were obtained, 250 of each of the three groups.

## 3. CONCLUSIONS

In this study, multifractal analyzes were performed using two programs FracLac and FracLab, three groups of tissue pictures: normal tissue, carcinoma, and adenomas. Then, statistical processing of the obtained results was made using the SPSS statistical treatment program, which is usually used in clinical trials. In this way, the answers to the hypotheses posed in this paper are obtained. The general conclusion is that the basis of the general hypothesis proved to be correct, that the parameters of multifractal analysis differ significantly for all three groups of tissue tissues observed, and therefore the zero hypothesis about the non-separation of these groups was denied. This applies when applying both programs for multifractal analysis of FracLac and FracLab images.

Since the general hypothesis of this paper is confirmed, it can be concluded that this research has obtained positive results.. In the case of the FracLac program, the reliability of the classification of all three tissue groups analyzed based on the obtained multifractal parameters is 65.3%, which is more than the 60.7% obtained in the case of FracLab. In the case of the FracLac program, 80.0%, 73.0% and 85.0% were obtained, successfully classified cases for the following group relationships (respectively): normal tissue and carcinomas, carcinomas and adenomas, normal tissue and adenomas.

In the case of the FracLab program for the same relationships, tissue image groups, respectively, received the following reliability of 64.0% (which is considerably worse in relation to the same case with the FracLac program), 74.0% (which is a little better in relation to the same case for the FracLac program) and 80.0% (which is worse in relation to the same case with FracLac), we can conclude that the resulting classifications are very effective and this is generally more in the case of FracLac.

**ACKNOWLEDGEMENTS**

This work was supported by the Ministry of Education and Science, Republic of Serbia, Science and Technological Development grant TR32037.